\documentclass[manuscript,sigconf,natbib=false]{acmart}

\AtBeginDocument{%
  }

\copyrightyear{2022} 
\acmYear{2022} 
\setcopyright{acmlicensed}
\acmConference[MMSports '22]{Proceedings of the 5th International ACM Workshop on Multimedia Content Analysis in Sports}{October 10, 2022}{Lisboa, Portugal}
\acmBooktitle{Proceedings of the 5th International ACM Workshop on Multimedia Content Analysis in Sports (MMSports '22), October 10, 2022, Lisboa, Portugal}
\acmPrice{15.00}
\acmDOI{10.1145/3552437.3555699}
\acmISBN{978-1-4503-9488-8/22/10}

\RequirePackage[
  datamodel=acmdatamodel,
  style=acmnumeric,
  ]{biblatex}

\usepackage{xspace} % handles space at the end of commands
\usepackage{multirow}

\begin{document}

\title{DeepSportradar-v1: Computer Vision Dataset for Sports Understanding with High Quality Annotations}

\author{Gabriel Van Zandycke}
\email{g.vanzandycke@sportradar.com}
\affiliation{\institution{Sportradar AG}\country{Switzerland}}
\orcid{0000-0002-8384-4166}
\authornote{Authors contributed equally to this work.}

\author{Vladimir Somers}
\email{v.somers@sportradar.com}
\affiliation{\institution{Sportradar AG}\country{Switzerland}}
\authornotemark[1]
\orcid{0000-0001-5787-4276}

\author{Maxime Istasse}
\email{m.istasse@sportradar.com}
\orcid{0000-0002-5153-1106}
\affiliation{\institution{Sportradar AG}\country{Switzerland}}
\authornotemark[1]

\author{Carlo Del Don}
\email{c.deldon@sportradar.com}
\orcid{0000-0001-9537-7078}
\affiliation{\institution{Sportradar AG}\country{Switzerland}}

\author{Davide Zambrano}
\authornotemark[1]
\email{d.zambrano@sportradar.com}
\orcid{0000-0003-3977-4647}
\affiliation{\institution{Sportradar AG}\country{Switzerland}}

\renewcommand{\shortauthors}{Gabriel Van Zandycke et al.}

\begin{abstract}
With the recent development of Deep Learning applied to Computer Vision, sport video understanding has gained a lot of attention, providing much richer information for both sport consumers and leagues. This paper introduces DeepSportradar-v1, a suite of computer vision tasks, datasets and benchmarks for automated sport understanding. The main purpose of this framework is to close the gap between academic research and real world settings. To this end, the datasets provide high-resolution raw images, camera parameters and high quality annotations. DeepSportradar currently supports four challenging tasks related to basketball: ball 3D localization, camera calibration, player instance segmentation and player re-identification. For each of the four tasks, a detailed description of the dataset, objective, performance metrics, and the proposed baseline method are provided. To encourage further research on advanced methods for sport understanding, a competition is organized as part of the MMSports\footnote{\url{http://mmsports.multimedia-computing.de/mmsports2022/index.html}} workshop from the ACM Multimedia 2022 conference, where participants have to develop state-of-the-art methods to solve the above tasks. The four datasets, development kits and baselines are publicly available\footnote{\url{https://github.com/DeepSportRadar}}.
\end{abstract}



%\settopmatter{printacmref=false}

\maketitle

\section{Introduction}

Individual and professional sports have always had a strong impact on the economic, political, and cultural aspects of our society. When only considering the economical side, this impact is likely going to increase as the global sports market size, including services and goods from sport entities, is expected to grow from $\$354.96$~billion in 2021 to $\$707.84$~billion in 2026\footnote{\url{https://www.thebusinessresearchcompany.com/report/sports-global-market-report}}. The online live-streaming market alone is going to increase in value from $\$18.12$bn in 2020 to $\$87.34$bn in 2028\footnote{\url{https://www. verifiedmarketresearch.com/product/sports-online-live-video-streaming-market}}.
A strong contribution to this growth is provided by the recent and rapid technological advancement which has changed the way people watch and enjoy sports. Indeed, Computer Vision (CV) and the recent developments in Deep Learning (DL)~\cite{lecun2015deep} provide the opportunity to extract meaningful information from live streamed events resulting in a much richer experience for both consumers and leagues.

The ability of DL-based solutions to be useful and reliable in real world applications strongly depends on the quantity and quality of data on which the model has been trained on in the first place~\cite{alom2019state}. Specifically for sports, the different disciplines and conditions make them unique in terms of the problems the model has to face. Moreover, the quality of the annotations often determines the model performances overall. In the past few years, the SoccerNet~\cite{giancola2018soccernet, deliege2021soccernet, cioppa2022scaling, cioppa2022soccernet} datasets have received increasing attention for the amount of data and the benchmark models provided to the CV community. However, two main issues pertain such initiative: first, considering only soccer as representative of all sports does not allow to extend results to other domains; secondly, and more importantly, the SoccerNet annotations are created out of broadcast videos which bring a series of concerns. These concerns include: a limited spatial and temporal coverage of the game due to, on one side, the frequent camera movements which return a subset of the field and, on the other, the replays or advertisements which interrupt the live stream; a lower image resolution with respect to the original sensor; no access to camera parameters or position; and the overlaying graphics such as scores, teams name, advertisements, game statistics, that obstruct the image. In summary, broadcast video annotations remain distant from the actual sensors and tools used to record the game. In conclusion, while initiatives as SoccerNet represent a strong and valid tool for the computer vision community, the introduction of an high quality dataset with available raw images and camera parameters, will help closing the gap between academic research and real world settings.

This paper introduces two datasets in the basketball domain and four different CV tasks each associated with a task-specific dataset extracted from the first two. The data and annotations are provided by SynergySports\footnote{\url{https://synergysports.com/}}, a division of Sportradar\footnote{\url{https://sportradar.com/}}, and have been recorded with the Keemotion/Synergy Automated Camera System™. The proposed four tasks are:
\begin{itemize}
    \item \textbf{Ball 3D localization in calibrated scenes.} This task tackles the estimation of ball size on basketball scenes given the oracle ball position. 
    \item \textbf{Camera calibration.} This task aims at predicting the camera calibration parameters from images taken from basketball games.
    \item \textbf{Player instance segmentation.} This task deals with the segmentation of individual humans (players, coaches and referees) on the basketball court.
    \item \textbf{Player re-identification.} In this task, the objective is to re-identify basketball players across multiple video frames captured from the same camera viewpoint at various time instants.
\end{itemize}

Moreover, a competition around the four tasks has been
organized and results will be presented at the 5th International ACM Workshop on Multimedia Content Analysis in Sports\footnote{\url{http://mmsports.multimedia-computing.de/mmsports2022/index.html}}.

A toolkit is provided for each task containing data, annotations and metrics. Moreover, a baseline for each task has been added which serves the purpose of providing an example to consume the data, and as a starting point for the challenge participants' solutions. The next section explains the original datasets, while the subsequent sections will describe each task in detail.

% image based challenges

\section{Datasets}
\label{sec:datasets}

The four tasks are built on two different datasets. The DeepSport dataset---a multi-labels dataset containing ball 3D annotations, image calibration data and human segmentation masks---is used for the ball 3D localization, court calibration and instance segmentation tasks. 
The DeepSportradar player ReID dataset is %, introduced in~\cite{vipriors},
used for the players re-identification task only and will be described further in Section \ref{sec:reid}.
%and described in section~\ref{sec:reid}.
Both datasets were acquired during professional basketball matches with the Keemotion/Synergy Automated Camera System™ that offers a sideline view of the court from a position aligned with the court center line. Images were fully annotated and reviewed by human operators, leading to high quality labels, and are made freely available to the research community by Sportradar.
% and the LNB (ligue nationale de basketball) of France accepted to release the right on the images for research purpose…

\subsection*{DeepSport dataset}
\label{sec:deepsport-dataset}
Hereafter, the multi-labels DeepSport dataset, used for three tasks, is described. Originally introduced in~\cite{ballseg} with only ball annotations, it was later supplemented with new data and additional annotations. It is now made available publicly on the Kaggle platform~\cite{kaggle-deepsport}.
%It is a multi-labels dataset containing ball 3D annotations, image calibration data and human segmentation masks, and was used to extract %build the task-specific dataset.

\begin{figure*}[h]
    \centering
    \includegraphics[width=\textwidth]{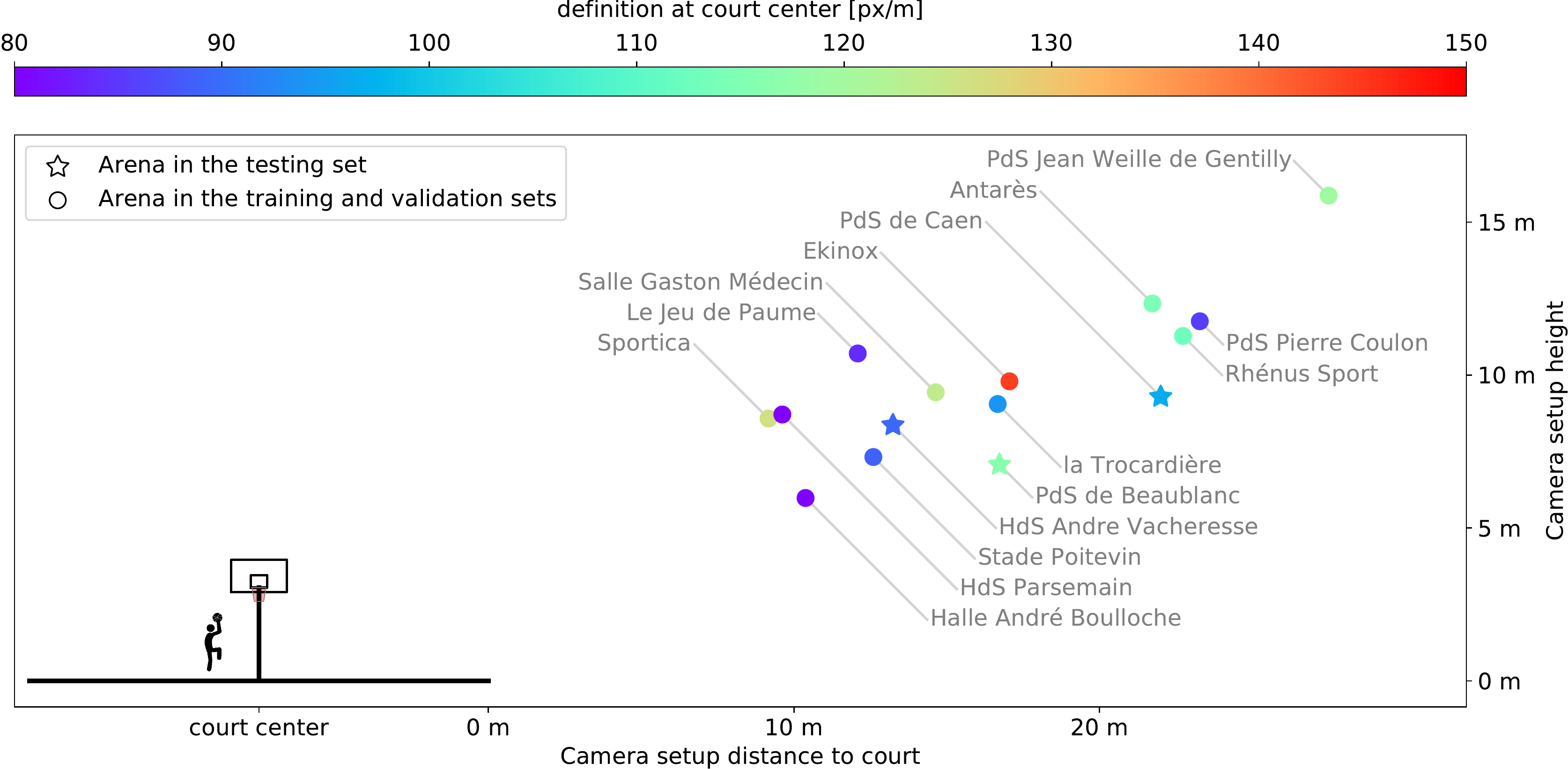}
    \caption{Cross section showing camera setup height from the ground and distance to the court of the different arenas in which images were acquired.
    The camera definition depends on camera resolution, sensor size, lens focal length and camera setup distance to the court.}
    \label{fig:crosssection}
\end{figure*}

\begin{figure}
    \centering
    \includegraphics[width=.49\columnwidth, trim=30 30 30 30, clip]{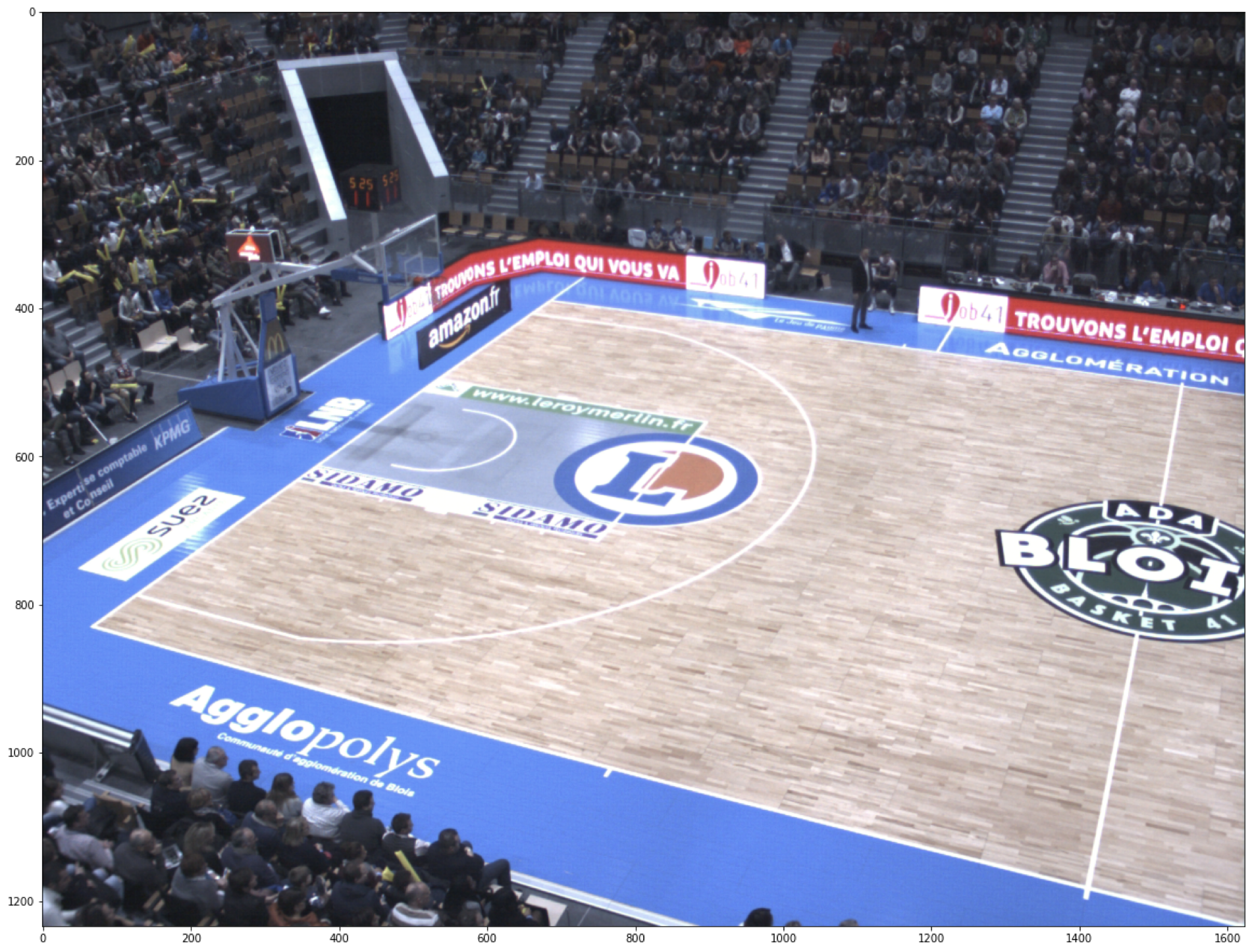}
    \includegraphics[width=.49\columnwidth, trim=30 30 30 30, clip]{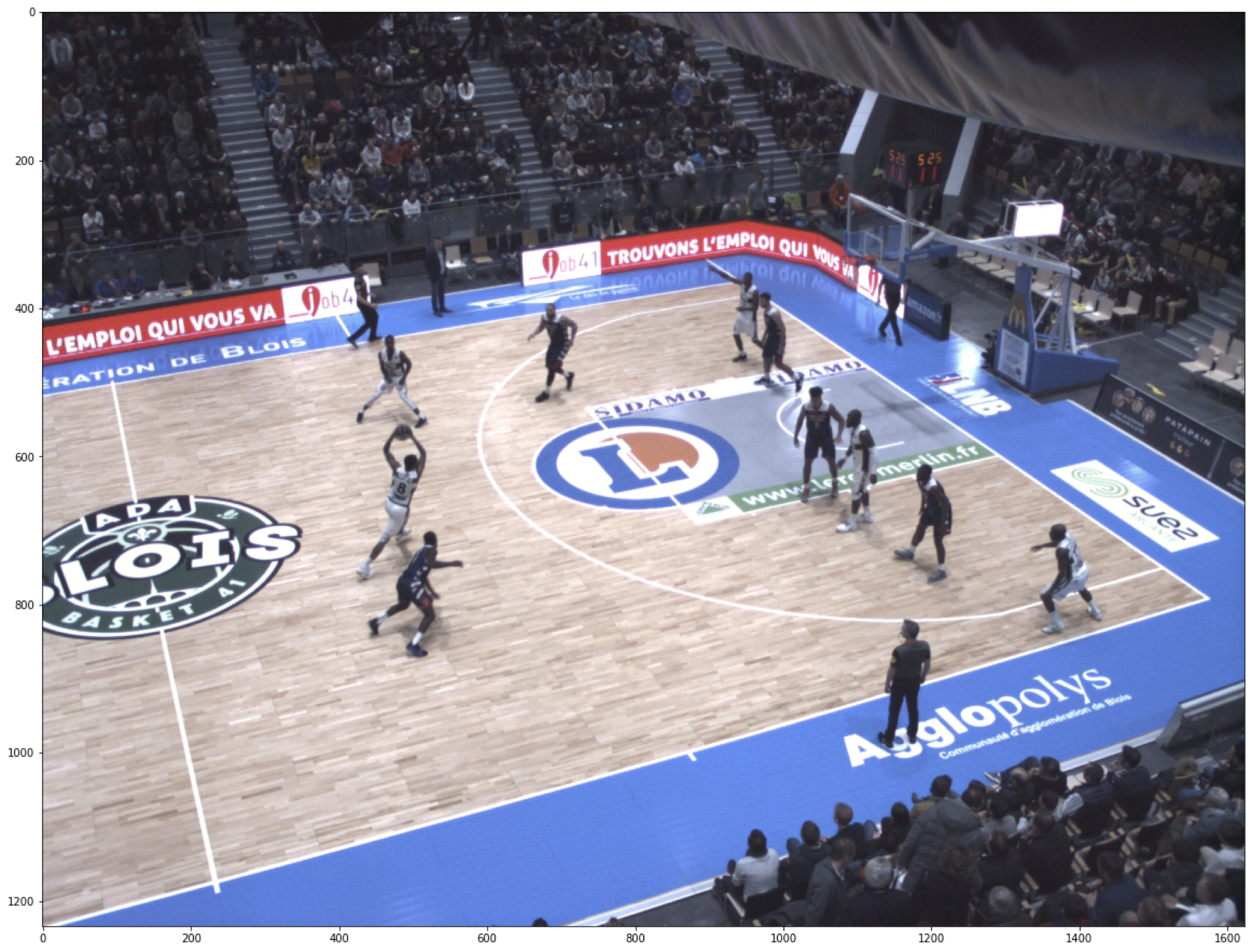}
    \caption{A \emph{raw instant} captured by the Keemotion/Synergy Automated Camera System with a two cameras setup.}
    \label{fig:pair}
\end{figure}

\paragraph{Description} 
The dataset is a collection of \emph{raw-instants}: sets of images captured at the same instant by an array of cameras covering a panorama of the sport field. It features only in-game basketball scenes.
Figure~\ref{fig:pair} shows a \emph{raw-instant} from a two cameras setup. In the DeepSport dataset, camera resolutions range from 2Mpx to 5Mpx. As illustrated in Figure~\ref{fig:crosssection}, the resulting images have a definition varying between 80px/m and 150px/m, depending on camera resolution, sensor size, lens focal-length and distance to the court. 

\paragraph{Origin}
The dataset was captured in 15 different basketball arenas, each identified by a unique label, during 37 professional games of the French league LNB-Pro~A.
Figure~\ref{fig:crosssection} depicts a cross section of a basketball court where the camera setup height and distance to the court is shown for each arena.

\newcommand{\testset}[1]{\textbf{#1}}

\begin{table}[tb]
    \caption{The DeepSport dataset was captured in 15 different arenas and three of them are kept for the testing set. It features a variety of angle of views, distance to the court and image resolution.}
    \label{tab:arena_sets}
    % \centering
    \resizebox{\columnwidth}{!}{%
    \begin{tabular}{@{}ll@{}c@{}}
\toprule
Arena label & Arena name (City) & \shortstack{Number \\ of items} \\
\midrule
\textsc{ks-fr-stchamond} & Halle André Boulloche\, (Saint-Chamond) & 12\\
\textsc{ks-fr-fos} & HdS Parsemain\, (Fos-sur-Mer) & 23\\
\textsc{ks-fr-strasbourg} & Rhénus Sport\, (Strasbourg) & 8\\
\textsc{ks-fr-vichy} & PdS Pierre Coulon\, (Vichy) & 9\\
\textsc{ks-fr-nantes} & la Trocardière\, (Nantes) & 20\\
\textsc{ks-fr-bourgeb} & Ekinox\, (Bourg-en-Bresse) & 12\\
\textsc{ks-fr-gravelines} & Sportica\, (Gravelines) & 129\\
\textsc{ks-fr-monaco} & Salle Gaston Médecin\, (Monaco) & 9\\
\textsc{ks-fr-poitiers} & Stade Poitevin\, (Poitiers) & 5\\
\textsc{ks-fr-nancy} & PdS Jean Weille de Gentilly\, (Nancy) & 40\\
\textsc{ks-fr-lemans} & Antarès\, (Le Mans) & 16\\
\textsc{ks-fr-blois} & Le Jeu de Paume\, (Blois) & 39\\
\testset{\textsc{ks-fr-caen}} & \testset{PdS de Caen\, (Caen)} & \testset{31}\\
\testset{\textsc{ks-fr-roanne}} & \testset{HdS Andre Vacheresse\, (Roanne)} & \testset{3}\\
\testset{\textsc{ks-fr-limoges}} & \testset{PdS de Beaublanc\, (Limoges)} & \testset{8}\\
\bottomrule
    \end{tabular}
    }
\end{table}

\paragraph{Split}
The dataset is split in 3 subsets: training, validation and testing. The testing set contains all images from arena labels \textsc{ks-fr-caen}, \textsc{ks-fr-limoges} and \textsc{ks-fr-roanne}. If not otherwise specified, the remaining instants are split in 15\% for the validation set and 85\% for the training set. A mapping between arena labels and arena names is given in Table~\ref{tab:arena_sets} and provides the amount of instants for each arena.
An additional challenge set, introduced for the competition, is composed of 35 additional similar instants. They come from a new set of arenas and labels will remain secret.
%An additional set of 35 similar instants was introduced for the competition. The labels of this challenge set will remain secret.

\paragraph{Annotations}
The cameras used to capture the \emph{raw-instants} are calibrated, which means that intrinsic and extrinsic parameters are known. The ball 3D annotation was obtained by leveraging the calibration data and clicking two points in the image space: the ball center and its vertical projection to the ground. This process is described and validated in~\cite{ball3d}. The contouring of humans lying near the court were annotated on each image individually following \cite{niels2022}, with a special care given to occlusions.

\section{The tasks}

This section describes the four tasks, explaining the dataset splits, the metrics used for evaluating and compare results for the competition and the baselines provided to the users.
\subsection{Ball 3D localization}

% TODO: what is unique and new about this task

%Estimating the ball 3D localization from a single calibrated image require more than detecting the ball in the image space.

Automated ball 3D localization in team sport scenes has important applications like assisting referees or feeding game analytics.
% In sports like basketball where the ball is often occluded and in players hands, the traditional multi-view ballistic trajectories approaches are limited.
In sports like basketball where the ball is often occluded and in players hands, an image based approach is required to fill the gap between the trajectories proposed by a ballistic approach.
Hence, this task aims at localizing the ball in 3D on basketball scenes, from a single calibrated image.
% to the best of our knowledge has only been studied in ~\ref{ball3d}
This problem can be solved by both detecting the ball center and estimating its size in the image space. Indeed, the 3D localization can be recovered using camera calibration information and the knowledge of the real ball size~\cite{ball3d}.
Since ball detection has been largely studied~\cite{kamble2019ball}, this task focuses on 3D localization given oracle ball positions. Hence, the task consists in estimating the ball diameter in pixels in the image space from an image patch centered on the ball.

% This challenge tackles the estimation of ball size on basketball scenes given oracle ball position. Using camera calibration information and knowledge of the real ball size, this estimation can be used to recover the ball 3d localization in the scene\cite{vanzandycke2022}.

\subsubsection{Dataset}

\begin{figure*}[ht]
\centering
\begin{tabular}{@{}c@{\;}c@{\;}c@{\;}c@{\;}c@{\;}c@{\;}c@{\;}c@{}}
\includegraphics[width=2cm]{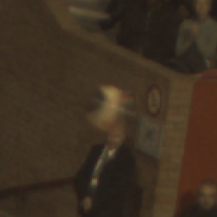}&
\includegraphics[width=2cm]{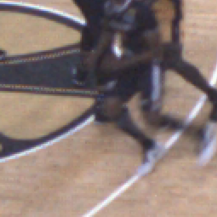}&
\includegraphics[width=2cm]{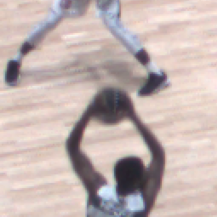}&
\includegraphics[width=2cm]{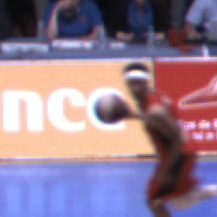}&
\includegraphics[width=2cm]{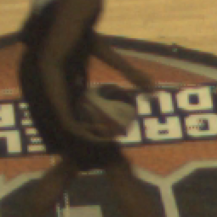}&
\includegraphics[width=2cm]{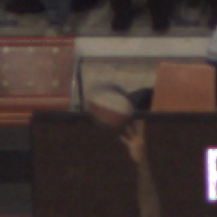}&
\includegraphics[width=2cm]{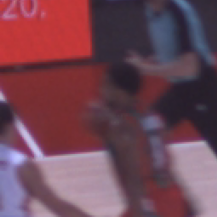}&
\includegraphics[width=2cm]{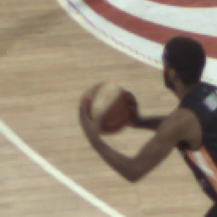}\\
\includegraphics[width=2cm]{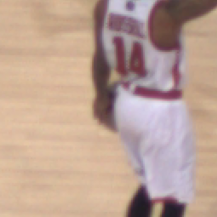}&
\includegraphics[width=2cm]{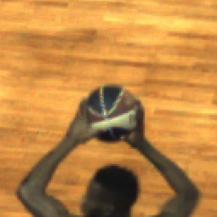}&
\includegraphics[width=2cm]{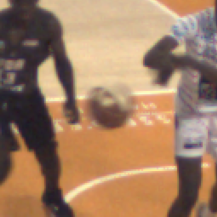}&
\includegraphics[width=2cm]{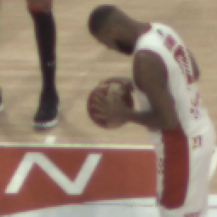}&
\includegraphics[width=2cm]{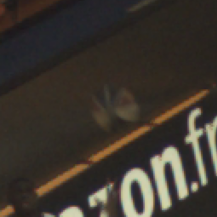}&
\includegraphics[width=2cm]{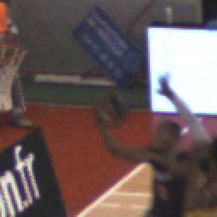}&
\includegraphics[width=2cm]{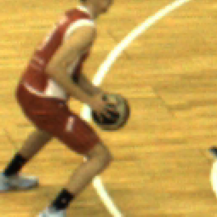}&
\includegraphics[width=2cm]{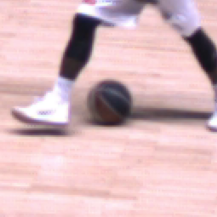}\\
\includegraphics[width=2cm]{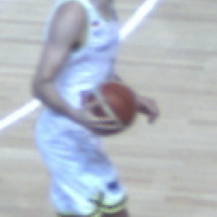}&
\includegraphics[width=2cm]{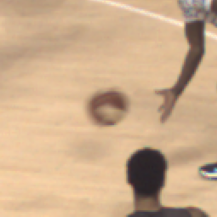}&
\includegraphics[width=2cm]{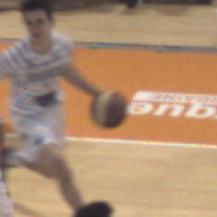}&
\includegraphics[width=2cm]{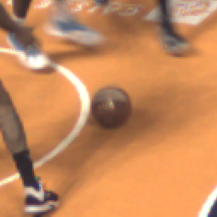}&
\includegraphics[width=2cm]{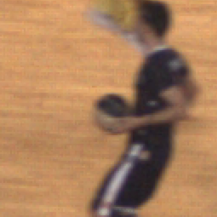}&
\includegraphics[width=2cm]{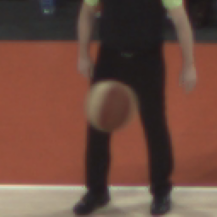}&
\includegraphics[width=2cm]{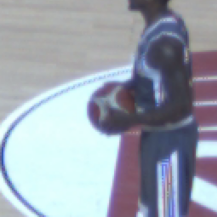}&
\includegraphics[width=2cm]{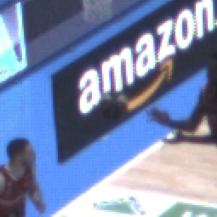}\\
\includegraphics[width=2cm]{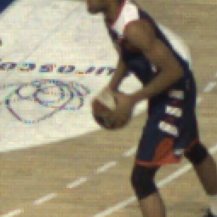}&
\includegraphics[width=2cm]{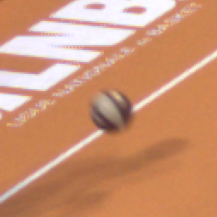}&
\includegraphics[width=2cm]{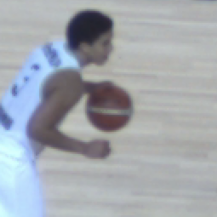}&
\includegraphics[width=2cm]{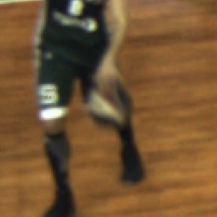}&
\includegraphics[width=2cm]{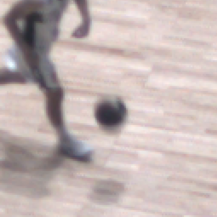}&
\includegraphics[width=2cm]{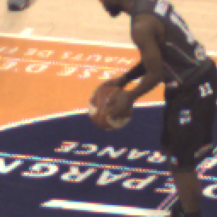}&
\includegraphics[width=2cm]{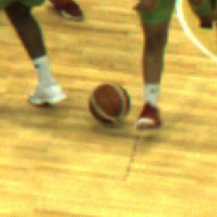}&
\includegraphics[width=2cm]{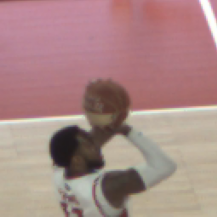}\\
\end{tabular}
    \caption{Samples of $128\times 128$ crops around oracle ball positions. The dataset features many different scenes with different colors, backgrounds and lighting conditions. Ball is often partly occluded or in players hand and suffers from motion blur.}
    \label{fig:ballsamples}
\end{figure*}

\begin{figure}[b]
    \centering
    \includegraphics[width=.6\columnwidth]{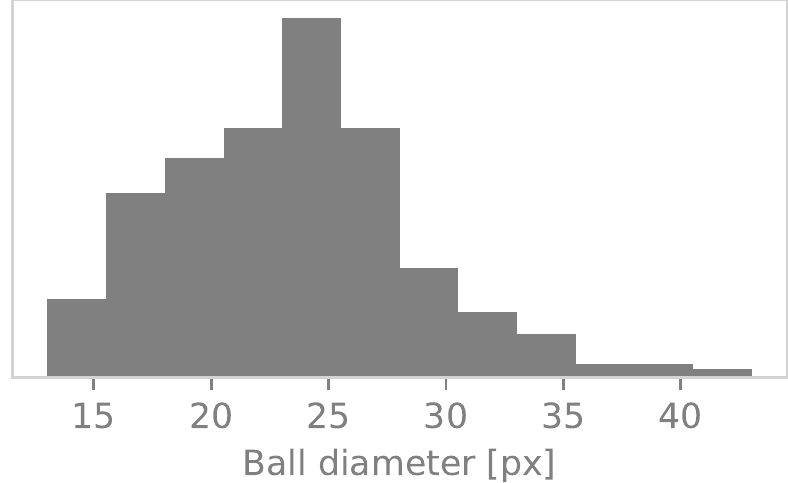}
    \caption{Ball size distribution in the DeepSport dataset.}
    \label{fig:balldistribution}
\end{figure}

The task uses $N\times N$ crops around oracle ball positions from the {\textbf{DeepSport dataset}} presented in Section~\ref{sec:deepsport-dataset}, where $N$ is a parameter. Figure~\ref{fig:ballsamples} shows samples from the dataset with $N=128$.
As shown in Figure~\ref{fig:balldistribution}, ball diameter ranges from 15px to 35px in the dataset.

\subsubsection{Metrics}

The main metric used to evaluate methods is the mean absolute diameter error (MADE) between the prediction and the ground-truth. In addition, the mean absolute projection error (MAPE) and the mean absolute relative error (MARE), described in~\cite{ball3d}, are used. The MAPE measures the error of a vertical projection of positions on the ground plane. The MARE measures the distance error relative to the camera position.

\subsubsection{Baseline and results}

The baseline proposed with this task is the regression model described in~\cite{ball3d}. It is composed of a VGG16~\cite{Simonyan2014} features extractor followed by 3 fully connected layers and is supervised with a Huber loss~\cite{Huber1964}.

The baseline is trained with random scaling and random color-gamma data augmentations on image patches with $N=64$. The supervision is performed using Adam optimizer for 100 epochs with an initial learning rate of $10^{-4}$, exponentially decayed by half during two epochs. This decay is applied every 10 epochs starting at epoch 50.

The baseline reaches a MADE of 2.12 pixels. This corresponds to a MAPE of 3.05 meters and a MARE of 10\%.

\subsection{Camera calibration}
The Camera calibration task aims at predicting the camera parameters from images taken from basketball games. The toolkit for this task can be found on the main DeepSportRadar GitHub page. 
Formally, this task objective is to predict the projection matrix, $P_{3\times 4}$ that maps a 3D point in homogeneous coordinates (a 4-dimensional vector) to a 2D point in homogeneous coordinates (3-dimensional vector) in the image space\footnote{\url{https://en.wikipedia.org/wiki/Camera_matrix}}. The projection matrix combines intrinsic (sensor and lens) and extrinsic (position and rotation) camera parameters as:
\begin{align}
    P :&= \left[\begin{matrix}K_{3\times 3};{\bf 0}_{3\times 1}\end{matrix}\right] \left[\begin{matrix}R_{3\times 3};T_{3\times 1}\\{\bf 0}_{1\times 3};1\end{matrix}\right] \nonumber \\
    &=K_{3\times 3}\left[\begin{matrix}R_{3\times 3};T_{3\times 1}\end{matrix}\right],
\end{align}
where $K$ is the matrix of intrinsic parameters, while $R, T$ are the rotation and translation matrices respectively\footnote{\url{https://ispgroupucl.github.io/calib3d/calib3d/calib.html}}. For Synergy/Keemotion produced images, the origin of the 3D world is located on the furthest left corner of the basketball court relative to the main camera setup; more precisely in the inner side of the court lines. The unit of length is the centimeter and axis orientation is given by $x$ along the court length, $y$ along the court width and $z$ pointing downward\footnote{\url{https://gitlab.com/deepsport/deepsport_utilities/-/blob/main/calibration.md}}. For simplicity, this task assumes that lenses have no distortion.

The camera calibration parameters are crucial for several CV tasks such as the 3D tracking of players in the field. These parameters can be retrieved on site; however, an automatic method that estimates them is needed when the field and the camera are not accessible anymore. 
The Camera calibration task falls under the sport-field registration tasks and takes advantage of the known official sizes of the basketball court\footnote{\url{https://en.wikipedia.org/wiki/Basketball_court}}. Several approaches have been adopted to solve sport-field registrations for different sport domains including tennis, volleyball and soccer~\cite{farin2003robust, yao2017fast}, relying essentially on keypoints retrieval methods. With the advent of Deep Learning, common approaches tackle the problem as a segmentation task~\cite{homayounfar2017sports, chen2019sports, sha2020end, cioppa2021camera}. The baseline introduced for this task adopts this approach.

\subsubsection{Dataset}
This task purpose is to predict the camera calibration parameters from a single frame of a basketball game. The dataset is made of 728 views (pairs of images and corresponding camera calibration parameters) randomly generated from the \textbf{DeepSport dataset}.
The random view generation process\footnote{See implementation at: \url{https://gitlab.com/deepsport/deepsport_utilities/-/blob/main/deepsport_utilities/ds/instants_dataset/views_transforms.py}} generates a random 3D position within the court and image limits on which the view will be centered. It then samples a pixel density between $\alpha\cdot 20$~px/m and $\alpha\cdot 60$~px/m (see Figure~\ref{fig:crosssection}), a rotation between $-10^{\circ}$ and $10^{\circ}$, and a boolean horizontal flip, to create a crop from the original image of size $\alpha\cdot 480\times \alpha\cdot 270$. 
Please note that, the test and challenge sets were provided with $\alpha=2$,
%In this task, alpha is set to $2$ for sets,
resulting in an output image dimension of $920\times 540$.
The corresponding affine transformation matrix is applied to the intrinsic camera matrix $K$  to produce the camera calibration parameters that correspond to the generated view.
%method samples uniformly a scale within two given arguments (defined in pixels per meters) then returns the crop of the original image and the corresponding camera calibration parameters
%Therefore, from the original images, a potentially infinite number of views can be generated.
%Note that during the views generation the lenses distortions are also removed to simplify the task.
%By default t
These views are divided in train (480), val (164) and test (84) sets. For this challenge, having a validation set on arenas not seen during training is of foremost importance, therefore the arenas of \textsc{ks-fr-nantes}, \textsc{ks-fr-blois} and \textsc{ks-fr-fos} are used for the validation set (see Table \ref{tab:arena_sets}). A final challenge split composed of 84 images is provided for the competition purpose and its camera parameters are kept secret. 

A few sport-field registration datasets have been publicly released so far among which the SoccerNet-v2 is the largest~\cite{deliege2021soccernet} (20028 images from 500 games). It is worth noticing that,
with our random view generation process, a potentially infinite number of views can be generated from the original dataset images.
%as already mentioned in the DeepSport dataset each view are generated randomly, providing, therefore, a potentially infinite number of views.

\begin{figure}[t]
\centering
\begin{tabular}{@{}c@{}c@{}}
\includegraphics[width=0.98\columnwidth]{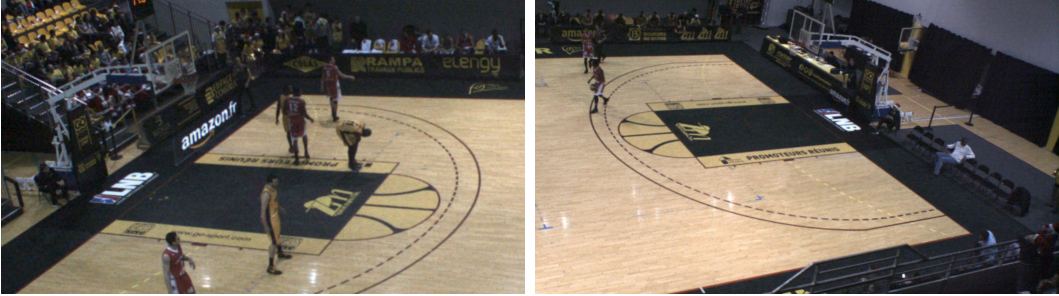}\\
\includegraphics[width=0.98\columnwidth]{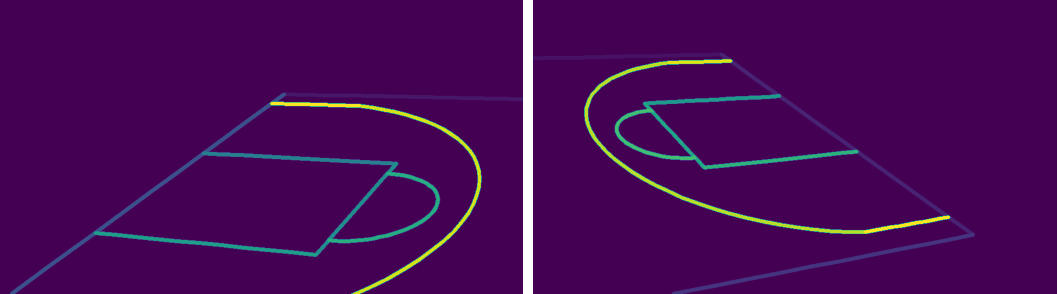}\\
\end{tabular}
    \caption{Examples of images from the camera calibration task (top). The target camera parameters have been used to generate the court lines (bottom), which can then be used as target for a segmentation model as described for the baseline.}
    \label{fig:cam_calib}
\end{figure}

\subsubsection{Metrics}
In order to evaluate the proposed methods on this task, the predictions are evaluated based on a Mean Squared Error (MSE) of the projection error of 6 points---left, center and right extremities at the middle and bottom parts of the frame---in the 3D coordinates.

\subsubsection{Baseline and results}
The baseline is composed by two models: the first is a segmentation model (DeepLabv3\cite{chen2017rethinking}) that predicts the 20 lines of the basketball court (see Figure \ref{fig:cam_calib}); the second finds the 2D intersections in the image space and matches them with the visible 3D locations of the court\footnote{\url{https://github.com/DeepSportRadar/camera-calibration-challenge/blob/main/utils/intersections.py}}. If enough intersections points are found (>5) the method \texttt{cv2.calibrateCamera}\footnote{\url{https://docs.opencv.org/4.6.0/d9/d0c/group__calib3d.html}} predicts the camera parameters. In all the other cases, the model returns an average of the camera parameters in the training set as default. The segmentation model has been fine-tuned on the Camera calibration dataset (with $\alpha=1$) for $40\text{k}$ steps with AdamW optimizer, base learning rate of $0.001$, weight decay of $0.0001$, and Amsgrad, reaching an mIoU of $0.46$ on the validation set.
The current baseline has an MSE error of $592.48$~cm on the Test split and $490.31$~cm on the Challenge set.
\subsection{Player instance segmentation}
The player instance segmentation task tackles the delineation of individual humans (players, coaches and referees) lying on a basketball court or less than 1 meter away from its borders.
Instance segmentation is a pervasive task that can apply to images captured from any domain in which objects can be individually identified and counted. Instance segmentation datasets have been collected, among other domains, in microbiology~\cite{kumar2019multi}, biology~\cite{minervini2016finely}, autonomous driving~\cite{cordts2016cityscapes} and in everyday life~\cite{gupta2019lvis,lin2014microsoft}.
The set of methods that solve instance segmentation is equally rich. A dichotomy is usually drawn between top-down and bottom-up methods. Top-down methods first propose candidate bounding boxes, and then segment the main object of interest in each of them~\cite{maskrcnn,bolya2019yolact}. Bottom-up methods first label pixels with embedding vectors and then cluster pixels with similar embeddings into instance masks~\cite{neven2019instance,cheng2020panoptic}.
In both types, main limitations usually arise in crowded regions, where objects are close to or occlude each other. So much so that many of the new state-of-the-art methods explicitly tackle those weaknesses~\cite{ke2021deep, yuan2021robust}.

\begin{figure*}[t]
\centering
\begin{tabular}{c@{\;}c}
\includegraphics[width=0.48\textwidth]{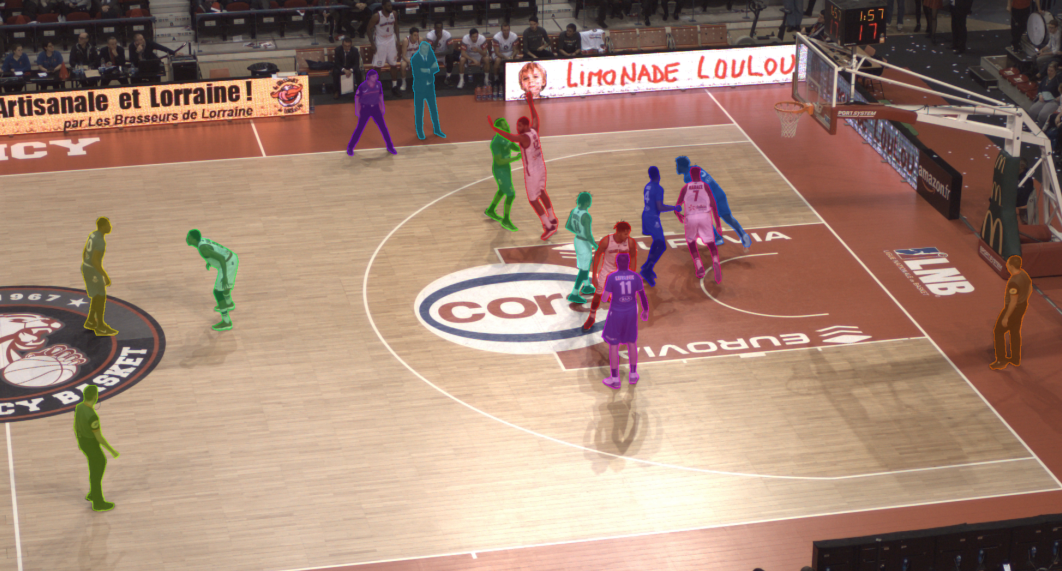}&
\includegraphics[width=0.48\textwidth]{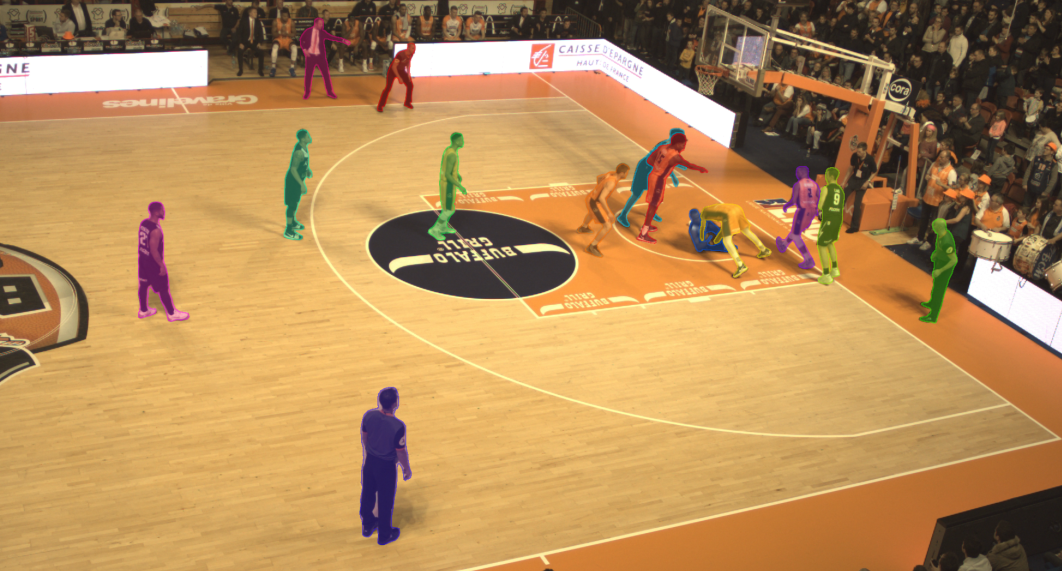}\\
\includegraphics[width=0.48\textwidth]{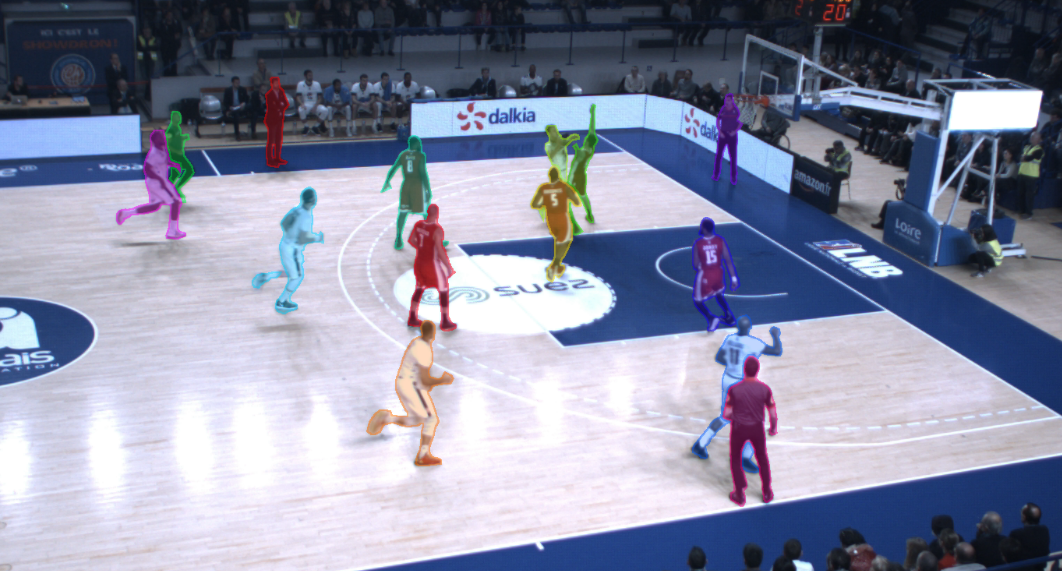}&
\includegraphics[width=0.48\textwidth]{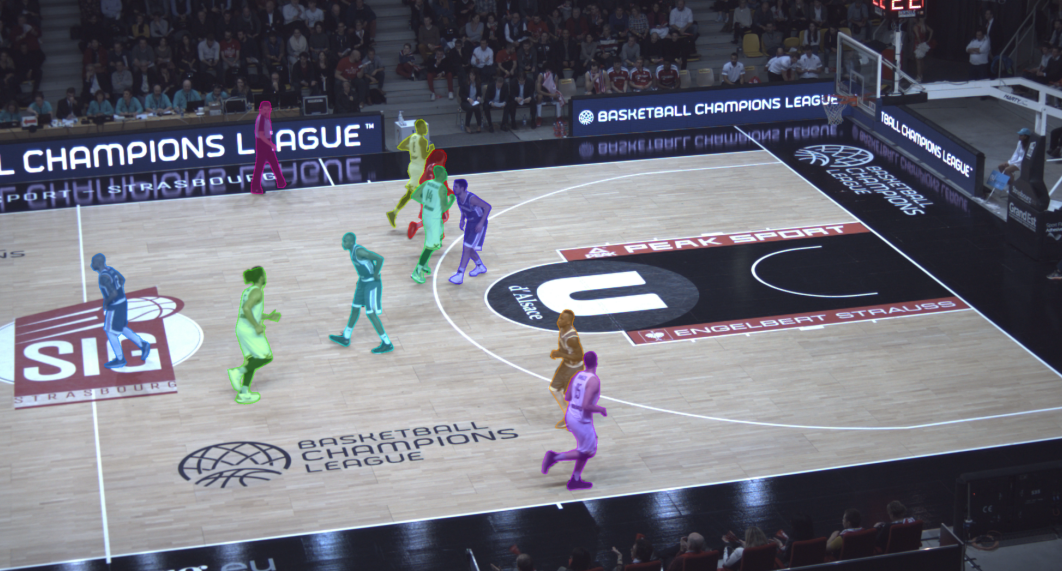}\\
\end{tabular}
    \caption{Samples of annotated images from the instance segmentation task (cropped around annotated instances). Annotated instances are highlighted in distinct colors.}
    \label{fig:segmentation_samples}
\end{figure*}

The dataset we propose here has three key aspects that make it particularly relevant for studying instance segmentation in those challenging cases. Please refer to Figure~\ref{fig:segmentation_samples} for visual examples that illustrate those.
First, instances only belong to one class. This renders the training and analysis of models less cumbersome, as there is no interference between classes of different frequencies during the training, and no averaging of performance metrics across them.
Second, although only one class is present, instances have varied appearances and poses, and are sometimes already tricky to extract from the background. Furthermore, occlusions are frequent, constituting challenging cases. The fact that instances of a same class have high interactions, sometimes leading each other to be split in disconnected parts, stresses greatly current instance segmentation methods.
Third, instance masks provided are very precise. Those annotations have been semi-automatically annotated as reported in~\cite{niels2022}.
All in all, we believe this dataset provides a good compromise that is challenging for state-of-the-art models, yet practical to study models with.

\subsubsection{Dataset}
The task uses the \textbf{DeepSport dataset} presented in Section~\ref{sec:deepsport-dataset}, with every images used individually\footnote{Recall that dataset items, the \emph{raw-instants}, are composed of multiple images}, and provided in the COCO format~\cite{lin2014microsoft}. The \textit{train} and \textit{val} subsets contain respectively 223 and 37 images 
sampled uniformly from the first set of arenas presented in Table~\ref{tab:arena_sets}. They contain respectively 1674 and 344 annotations. The \textit{test} subset contains 64 images coming from the last three arenas of Table~\ref{tab:arena_sets}. It contains 477 annotated humans.
%A total of 324 pictures from in-game instants are provided with high-quality contouring of humans lying near the court. The dataset is provided in the COCO format \cite{lin2014microsoft}. Three default splits are proposed. The \textit{train} and \textit{val} splits contain respectively 223 and 37 images sampled uniformly from the first set of arenas presented in Table~\ref{tab:arena_sets}, comprising respectively 1674 and 344 annotations. The \textit{test} split comes from the smaller set of arenas, comprising 64 images with 477 annotated humans. (in blue in Table~\ref{tab:arena_sets})
For the competition, participants were evaluated on the 84 images coming from the challenge set introduced in Section~\ref{sec:deepsport-dataset}. The number of annotated humans is kept secret.

%Competitors' submissions are evaluated on an additional subset of 84 images from other arenas. Its annotations are kept secret.

\subsubsection{Metrics}

Because the mAP metric is well established in instance segmentation~\cite{lin2014microsoft,cordts2016cityscapes}, and because we are mostly interested about segmentation quality, we focus on it as main metric. The version specific to instance segmentation (sometimes referred to as segm\_mAP) is different from that of object detection (bbox\_mAP). Indeed, it uses the intersection-over-union (IoU) of predicted and ground-truth masks rather than bounding boxes to compute each intermediate \rm{AP} curve. This way, good segmentation ($\rm{IoU} \ge 0.80$) is strongly rewarded while low-quality segmentation ($\rm{IoU} \approx 0.55$) is not.

Because only one class is present, there is no averaging between the metrics of frequent and rare classes. Our \rm{mAP} simply looks like
\begin{equation}
\rm{mAP} = \frac{1}{10}\sum_{\tau\in[0.50:0.05:0.95]} \rm{AP}@\tau
\end{equation}
$\rm{AP}@\tau$ being the area under the precision-recall curve, when considering as true-positives the predicted masks that have $\rm{IoU}>\tau$ with any ground-truth mask. ($\tau \ge 0.5$ prevents one-to-many mappings)

%$$\texttt{AP}@\tau = \int_0^1 p_\tau(r)dr$$
%$$
%p_\tau(r) = \frac{|\rm{TP} \text{ among } \mathcal{I}_r|}{r|GT|}
%$$
%$\mathcal I_r$ being the set of most confident predictions comprising $r|\rm{GT}|$ true instances.

\subsubsection{Baseline and results}

The code base we provide\footnote{\url{https://github.com/DeepSportRadar/instance-segmentation-challenge}} is built on MMDet \cite{mmdetection}. Our baseline consists of a Mask-RCNN model \cite{maskrcnn}, with a ResNeXt 101 backbone \cite{xie2017aggregated} and default configuration, trained for 20 epochs. This model reaches a mAP of $0.51$.

Relying on MMDet gives contestants the possibility to use a wide range of renowned and top-performing models off-the-shelf.

\subsection{Player re-identification}
\label{sec:reid}

\begin{figure}
    \centering
    \includegraphics[width=\columnwidth]{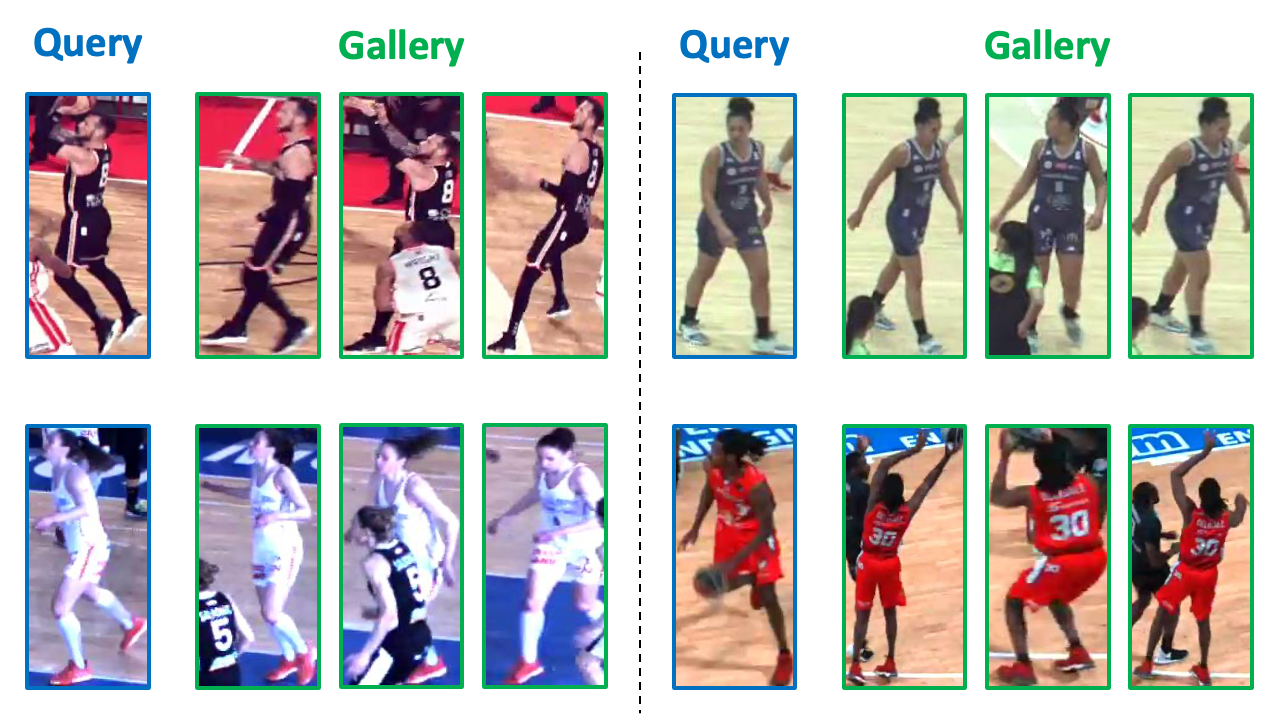}
    \caption{The player re-identification task: illustration of some correctly retrieved gallery samples for four players of interest given as queries.}
    \label{fig:player_reid_samples}
\end{figure}

Person re-identification \cite{Ye2022DeepLF}, or simply ReID, is a person retrieval task which aims at matching an image of a person-of-interest, called \textit{the query}, with other person images within a large database, called \textit{the gallery}, captured from various camera viewpoints. 
ReID has important applications in smart cities, video-surveillance and sports analytics, where it is used to perform person retrieval or tracking. 
The objective of the DeepSportradar player ReID task is to re-identify players, coaches and referees across images captured successively from the same moving camera during a basketball game, as illustrated in Figure \ref{fig:player_reid_samples}.
% In this challenge, participants will have to re-identify basketball players across multiple video frames captured from the same camera viewpoint at various time instants.
Compared to traditional street surveillance type re-identification datasets \cite{Miao2019PoseGuidedFA, Zheng2015ScalablePR, Zheng2017UnlabeledSG, li2014deepreid, Zheng2016MARSAV}, the DeepSportradar ReID dataset is challenging because players from the same team have very similar appearance, which makes it hard to tell them apart.
However, as opposed to standard ReID datasets, all images are captured by the same camera, from the same point of view.

ReID has gained more and more attention recently, with several works proposing state-of-the-art methods based on global \cite{fu2020unsupervised, he2020fastreid, Luo2019BagOT, Wang2018LearningDF}, or part-based \cite{Sun2018BeyondPM, Li2021DiversePD} feature extractor.
Other works introduced alternative ReID tasks, such as occluded ReID \cite{Miao2019PoseGuidedFA} or video-based ReID \cite{Zheng2016MARSAV}.
Multiple frameworks were also open-sourced to support further research on supervised \cite{torchreid, he2020fastreid} or unsupervised ReID \cite{ge2020selfpaced}.

% For this reason, the DeepSportradar player ReID task is more similar to a tracking objective than a multi-camera person retrieval objective.

\begin{table}[t]
\caption{The player re-identification dataset in numbers.}
\label{tab:reid_dataset_stats}
\centering
%\begin{tabular}{l|c|c|c|c}
% Title
%\textbf{Subset} & \textbf{\# Sequences} & \textbf{Split} & \textbf{\# Ids} & \textbf{\# Images}\\ 
%\midrule
% Content
%\multirow{1}{*}{Train} & \multirow{1}{*}{45} & - & 436 & 8569 \\
%\hline
%\multirow{2}{*}{Test} & \multirow{2}{*}{5} 
% & Query & 50 & 50 \\
% & & Gallery & 50 & 910 \\
%\hline
%\multirow{2}{*}{Challenge} & \multirow{2}{*}{49}
% & Query & - & 468 \\
% & & Gallery & - & 8703 \\
%\hline
%\end{tabular}
\begin{tabular}{l@{}clc@{}c}
% Title
\toprule
\textbf{Subset} & \textbf{\# Sequences} & \textbf{Split} & \textbf{\# Ids} & \textbf{\# Images}\\ 
\toprule
% Content
\multirow{1}{*}{Train} & \multirow{1}{*}{45} & - & 436 & 8569 \\
\midrule
\multirow{2}{*}{Test} & \multirow{2}{*}{5}  & Query & 50 & 50 \\
\cmidrule(lr){3-5}
 & & Gallery & 50 & 910 \\
\midrule
\multirow{2}{*}{Challenge} & \multirow{2}{*}{49} & Query & {\color{gray}(undisclosed)} & 468 \\
\cmidrule(lr){3-5}
 & & Gallery & {\color{gray}(undisclosed)} & 8703 \\
\bottomrule

\end{tabular}
\end{table}

% SynergyReID dataset loaded
%   subset      | # ids | # images
%   ---------------------------
%   train       |   436 |     8569
%   query test   |    50 |       50
%   gallery test |    50 |      910
%   traintest    |   486 |     9529
%   ---------------------------
%   query challenge  |   ? |      468
%   gallery challenge |  ? |     8703

% items 99 -> 
% arenas 29
% games 97
% was originally introduced for the VIPriors 2021 workshop~\cite{vipriors}.
\subsubsection{Dataset}
The DeepSportradar player ReID dataset was built using 99 short video sequences from 97 different professional basketball games of the LNB~proA league played in 29 different arenas.
This dataset of short basketball video sequences was originally introduced for the VIPriors 2021 workshop~\cite{vipriors}, and is different from the DeepSport dataset described in Section \ref{sec:datasets}.
These sequences are 20 frames long, with a frequency of 10 frames per second (FPS), and they contain on average 10 different tracklets, i.e. identities.
% These sequences comprises various players/sportswear appearance and have been recorded for various teams, jersey colors and arenas, 
Therefore, the dataset contains a wide variety of players and sportswear appearance, within multiple arenas with different illumination and court appearance.
Image crops\footnote{We refer to these image crops as "player thumbnails" for conciseness, without mentioning coaches and referees, because the large majority of these thumbnails actually depicts players.} from players, coaches and referees have been extracted within each of these frames.
The resulting re-identification dataset is composed of 18.703 thumbnails divided into three subsets: train, test and challenge set, as summarized in Table \ref{tab:reid_dataset_stats}.
Similar to other ReID datasets, the subset used for evaluating model performance, namely the test and challenge set, are split into a query and a gallery set.
For these sets, we chose thumbnails from the $1^{st}$ frame of the sequence as queries, and remaining thumbnails from the $2^{nd}$ to the $20^{th}$ frame as galleries.
Labels from the challenge are kept secret to avoid any cheating in the DeepSportradar Player ReID challenge.

\subsubsection{Metrics}
Two standard retrieval evaluation metrics are used to compare different ReID models:  the mean average precision \cite{Zheng2015ScalablePR} (mAP), and the cumulative matching characteristics (CMC) \cite{Wang2007ShapeAA} at Rank-1 and Rank-5.
The mAP is used to assess the average retrieval performance considering all ranked gallery samples.
The Rank-K accuracy is the probability that at least one correct match appears in the top-K ranked retrieved results.
Participants to the DeepSportradar Player ReID challenge are ranked according to their mAP score on the challenge set.

\subsubsection{Baseline and results}
% provide baseline on open-reid
% open reid is
% for the baseline, we rely on a standard architecture to wolve the reid task + ref
% backbone
% loss
% batch size, dropout, feature size, images size, optimizer, dist metric
% reference to more advanced methods

Person re-identification is generally formulated as a metric learning task \cite{Ye2022DeepLF}.
Firstly, a feature vector, also called "embedding", is extracted for each image in the dataset using a feature extractor.
Secondly, the query to gallery similarity scores are measured as the pairwise euclidean distance of these features vectors in the embedding space.
% \paragraph{Implementation details}
To address the DeepSportradar ReID challenge, we provide a simple CNN-based feature extractor as a baseline.
This feature extractor was implemented using the Open-ReID\footnote{https://github.com/Cysu/open-reid} framework, a lightweight library of person re-identification, open-sourced for research purpose.
Open-ReID aims to provide a uniform interface for different datasets, a full set of models and evaluation metrics.
The baseline employed a ResNet-50 \cite{He2016DeepRL} CNN as backbone and is trained with a classification objective: the model tries to predict each sample identity among the 436 identities in the training set.
The model is trained for 50 epochs with an SGD optimizer and a cross-entropy loss.
Training batches are made of 64 players thumbnails, all resized to $256\times128$.
We refer readers to our open-source toolbox on GitHub\footnote{\url{https://github.com/DeepSportRadar/player-reidentification-challenge}} for more details about the baseline architecture and training setup.

% \paragraph{Results}
The baseline achieves \textbf{65\% mAP}, \textbf{90\% Rank-1} and \textbf{96\% Rank-5} on the testing set of the DeepSportradar Player ReID dataset.

% future work/improvement: use more advanced methods and backbones, apply state-of-the-art methods from the litterature, BoT, fastreid, exploit specificity of challenge: same camera, thumbnails comes from successive frames from a video sequence with high frame rate. 

\section{Conclusions}
This paper has introduced two new datasets, namely the DeepSport dataset and the Basketball ReID dataset both acquired during professional basketball games with the Keemotion/Synergy Automated Camera System™. Together with these datasets, four CV tasks have been set up: the Ball 3D localization, Camera calibration, Player instance segmentation and Player re-identification. For each task, the dataset, the metrics and the baseline have been specified. The aim of this contribution was to provide a high-quality sports dataset framework where images, camera parameters and annotations are available and built close to the actual game recording setup, therefore providing an unique tool to experiment methods and solutions on real world settings.

\begin{acks}
We would like to thank the LNB (Ligue Nationale de Basketball) of France for allowing us to use and share images from their matches for the benefit of the research community.
\end{acks}

% Identification of funding sources and other support, and thanks to
% individuals and groups that assisted in the research and the
% preparation of the work should be included in an acknowledgment
% section, which is placed just before the reference section in your
% document.

% This section has a special environment:

% so that the information contained therein can be more easily collected
% during the article metadata extraction phase, and to ensure
% consistency in the spelling of the section heading.

% Authors should not prepare this section as a numbered or unnumbered {\verb|\section|}; please use the ``{\verb|acks|}'' environment.

% \section{Appendices}

% If your work needs an appendix, add it before the
% ``\verb|\end{document}|'' command at the conclusion of your source
% document.

% Start the appendix with the ``\verb|appendix|'' command:
% \begin{verbatim}
%   \appendix
% \end{verbatim}
% and note that in the appendix, sections are lettered, not
% numbered. This document has two appendices, demonstrating the section
% and subsection identification method.

% %%
% %% The acknowledgments section is defined using the "acks" environment
% %% (and NOT an unnumbered section). This ensures the proper
% %% identification of the section in the article metadata, and the
% %% consistent spelling of the heading.
% \begin{acks}
% \end{acks}

% %%
% %% Print the bibliography
% %%
\printbibliography

% %%
% %% If your work has an appendix, this is the place to put it.
% \appendix

\end{document}